\documentclass[runningheads]{llncs}
\usepackage{graphicx}
\usepackage{comment}
\usepackage{amsmath,amssymb} 
\usepackage{color}
\usepackage{mathrsfs}
\usepackage{verbatim}
\usepackage{diagbox}
\usepackage{multirow}
\usepackage{booktabs}
\usepackage{makecell}
\usepackage{authblk}
\usepackage{hyperref}

\makeatletter
\newcommand{\printfnsymbol}[1]{%
  \textsuperscript{\@fnsymbol{#1}}%
}
\makeatother
\newcommand{\samelineand}{\quad\quad}

\usepackage{geometry}
\begin{document}
\pagestyle{headings}
\mainmatter
\def\ECCVSubNumber{3268}  

\title{Peeking into occluded joints: A novel framework for crowd pose estimation} 




\titlerunning{Peeking into occluded joints}
%

\author{Lingteng Qiu\thanks{ indicates equal contributions}\inst{1,2,3} \and
Xuanye Zhang\printfnsymbol{1}\inst{1,2} \and
Yanran Li\inst{4} \and Guanbin Li\inst{5} \and Xiaojun Wu\inst{3} \and Zixiang Xiong\inst{6} \and Xiaoguang Han\thanks{corresponding author, hanxiaoguang@cuhk.edu.cn}\inst{1,2} \and Shuguang Cui\inst{1,2}}
\authorrunning{Lingteng et al.}
%
\institute{The Chinese University of Hong Kong, Shenzhen$^1$ \samelineand 
Shenzhen Research Institute of Big Data$^2$ \samelineand Harbin Institute of Technology(Shenzhen)$^3$ \samelineand Bournemouth University$^4$ \samelineand Sun Yat-sen University$^5$ \samelineand Texas A\& M University$^6$}
\maketitle

\begin{abstract}
	Although occlusion widely exists in nature and remains a fundamental challenge for pose estimation, existing heatmap-based approaches suffer serious degradation on occlusions. Their intrinsic problem is that they directly localize the joints based on visual information; however, the invisible joints are lack of that. In contrast to localization, our framework estimates the invisible joints from an inference perspective by proposing an Image-Guided Progressive GCN module which provides a comprehensive understanding of both image context and pose structure. Moreover, existing benchmarks contain limited occlusions for evaluation. Therefore, we thoroughly pursue this problem and propose a novel OPEC-Net framework together with a new Occluded Pose (OCPose) dataset with 9k annotated images. Extensive quantitative and qualitative evaluations on benchmarks demonstrate that OPEC-Net achieves significant improvements over recent leading works. Notably, our OCPose is the most complex occlusion dataset with respect to average IoU between adjacent instances. Source code and OCPose will be publicly available.
\end{abstract}

\section{Introduction}

Human pose estimation is a long-standing problem in Computer Vision. It has still attracted increasing attentions in recent years due to rising demands for wide range of applications which require human pose as input \cite{bansal2018recycle,chan2018everybody,gui2018teaching,joo2018total,mehta2017vnect,panteleris2018using,qian2018pose}. Despite the significant progress achieved in this area by advanced deep learning techniques \cite{li2019crowdpose,fang2017rmpe,cao2017realtime,he2017mask,xiao2018simple}, pose estimation in crowd scenarios still remains extremely challenging due to the intractable occlusion problem.

\begin{figure}[h]
\label{front}
\begin{center}
   \includegraphics[width=1\linewidth]{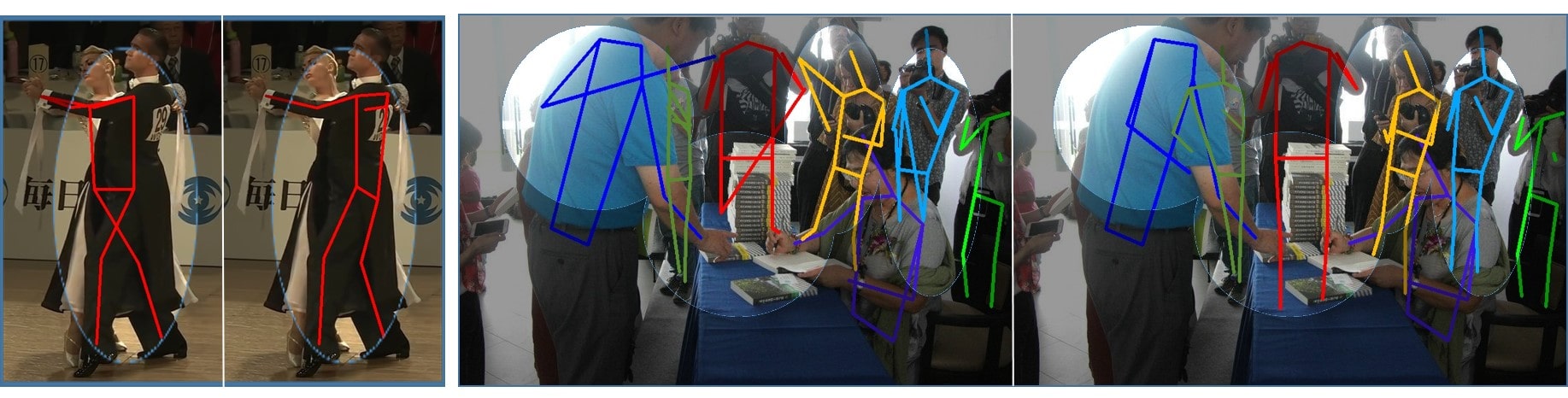}
\end{center}
 \caption{The current SOTA method \cite{li2019crowdpose} (left) VS our method (right).  Our method demonstrates a more natural and accurate estimation for occluded joints 
 }
\label{fig:FrontIMG}
\end{figure}

Trending models for crowd pose estimation strongly rely on heatmap representation for joints estimation: albeit being effective for visible joints, these methods still suffer performance degradation on occlusions and this is due to the fact that, since invisible joints are hidden, it is infeasible to directly localize them. To date, researchers have made painstaking efforts and complicated remedies in developing heatmap models and improving their accuracy of localization. However, the occlusion problem has only received little attention and only few attempts have been made into solving it. As illustrated in Fig \ref{front}, the current state-of-the-art work still produces very awkward poses and fails to estimate the occluded joints.

Occlusion is an intractable challenge in pose estimation due to the complicated background context, complex intertwined human poses and arbitrary occluded shape. To reveal the hidden joints, it becomes necessary to have a comprehensive inference method rather than simple localization. \textit{Our key insight is that the invisible joints are strongly related to contextual understanding of the image and structural understanding of the human pose.} For example, humans can easily infer the location of invisible joints using clues derived from the action type and the image context. Therefore, we delve deeper into the clues needed for invisible joints inference and propose a novel framework OPEC-Net to incorporate these clues for multi-person pose estimation.
To achieve this goal, two stages are proposed in our framework: Initial Pose Estimation and GCN-based Pose Correction. The first stage generates heatmaps to produce an initial pose and the subsequent correction stage adjusts the initial pose obtained from the heatmaps by an Image-Guided Progressive GCN (IGP-GCN) module.

The correction module deals with the image context and pose structure clues in the following aspects: (1) The human body structure provides the essential constraint information between joints. For this reason, the correction module is designed as a GCN-based network, which offers an explicit way of modeling the body structural information that is advantageous for correcting the joints. (2) Another important clue to infer the invisible joints is their related image context. Considering that, our GCN network is specially designed in an Image-Guided way: The IGP-GCN feeds both the coordinate of joints and also the image features extracted at the location of joints as input to each graph node. Therefore, the multi-scale image features from the heatmap modules are fed into the IGP-GCN in a progressive way, so that large displacements can be learned steadily. This enables the IGP-GCN to not only capture pose structural information but also the contextual image information at the same time. (3) In a crowd scenario, human interaction information becomes vital to infer poses. Therefore, we further formulate a CoupleGraph by connecting the corresponding joints of two instances, making the interaction between the pair of people contribute to our estimation results as well. 
However, the multi-scale image features learnt for heatmap estimation are not compatible for the coordinate correction module. Thus, a Cascaded Feature Adaption (CFA) strategy is introduced to process the features first: since the finer image feature has lost more global contextual information, we fuse the low-level features with high-level features following a cascaded design in order to strengthen their contextual information. 

Finally, our framework is trained in an end-to-end fashion and addresses occlusion problem in an elegant way. Interestingly, the heatmap module and coordinate GCN module are complementary in our framework: the quantisation error introduced from the heatmap modules can be addressed by the IGP-GCN and, at the same time, the heatmap modules offer a more accurate initial value for IGP-GCN that benefits the correction.

We conduct comprehensive experiments and introduce a new dataset to evaluate our framework. While occlusion cases are ubiquitous in crowded scenarios, only few existing benchmarks include enough complex examples tailored to the evaluation of this problem. Thus, it becomes necessary to have datasets that that not only contain light occlusions but also include heavily occluded scenes, such as waltz and wrestling, in which individuals are intertwined in complex ways. However, this field is still lacking such datasets because annotating human poses in heavily occluded scenes is very difficult and requires massive manual work. Therefore, we introduce a new dataset called \textbf{Occluded Pose (OCPose)} that includes more complex occlusions. We manually label all the 18k groundtruth human poses of the 9k images in OCPose. We also compare the average intersection over union (IoU) with typical datasets. MSCOCO \cite{lin2014microsoft} and MPII \cite{andriluka20142d} have less than 5\% data with IoU higher than 30\%, in contrast, our dataset OCPose contains 90\% data with IoU higher than 30\%. 

In summary, the contributions of this work are:
\begin{itemize}
    \item To the best of our knowledge, this is the first attempt at tackling the challenging problem of occluded joints from digging the image context and pose structure clues in an inference perspective. A novel framework, named OPEC-Net, is proposed, which significantly outperforms existing methods. 
    
    \item Our approach designs a novel Image-Guided Progressive GCN to accommodate the structural pose information and contextual image information for correction in a single pass. 
    
    \item We contribute a carefully-annotated 9K human pose dataset \textbf{OCPose} that includes highly challenging occluded scenes. To the best of our knowledge, OCPose is one of the datasets that contains the most complex occlusions to date. The OCPose dataset will be released to the public to facilitate research in the pose estimation field. 
\end{itemize}

\section{Related Works}
\textsl{\textbf{Heatmap-based Models for pose estimation.}}

Models for multi-person pose estimation (MPPE) can be divided into two categories, namely bottom-up and top-down approaches. The bottom-up methods first detect the joints and then assign them to the matching person. Pioneer works of bottom-up methods \cite{pishchulin2016deepcut,insafutdinov2016deepercut,cao2017realtime,newell2017associative,zanfir2018deep} attempted to design different joint grouping strategies. Newell \textit{et al. } \cite{newell2017associative} introduced a stacked hourglass network to utilize the tagging heatmap. DeepCut \cite{pishchulin2016deepcut} presented an Integer Linear Program (ILP) and Zanfir \textit{et al.} \cite{zanfir2018deep} grouped joints by learned scoring functions. Cao \textit{et al.} \cite{cao2017realtime} proposed a novel 2d vector field Part Affinity Fields (PAFs) for association as well. However, these prior works all have a serious deficiency that the invisible joints will decrease the performance drastically.

In the second category, the top-down methods first detect all people in the scene and then perform pose estimation for each person. Most of the existing top-down approaches \cite{he2017mask,papandreou2017towards,fang2017rmpe} focused on proposing a more effective human detector to obtain better results. Fang \textit{et al.} \cite{fang2017rmpe} proposed a framework which is more robust for the redundant human bounding box. Li \textit{et al.} \cite{li2019crowdpose} designed a global maximum joints association algorithm to address the association problem in crowd scenarios. Nevertheless, all of these strategies are unable to adequately reduce errors, especially in the severe occlusion cases, where one bounding box captures joints of multiple people. Most of the mainstream approaches are heatmap-based and thus are limited to estimating invisible joints which are lack of visual information. Therefore, we propose an OPEC-Net which completely differ from these works and is able to estimate invisible joints by inference rather than by localization.

\textsl{\textbf{GCN for pose modelling.}} 
The human body shows a natural graph structure, so that some advanced work constructed graph networks to address human pose related problems, such as action recognition \cite{yan2018spatial}, motion prediction \cite{li2019actional}, 3D pose regression \cite{zhao2019semantic,ci2019optimizing}. These work intuitively form the natural human pose as a graph and apply convolutional layers on it. Compared to other approaches, Graph Convolutional Networks demonstrate one compelling advantage when deal with human pose modeling problem: they are more effective in capturing dependency relationships between joints. 

Previous work \cite{yan2018spatial,li2019actional} achieved a significant gap of improvements in human motion understanding by forming the spatial and temporal relationships as edges in graph. Moreover, pose regression from 2D to 3D is a natural graph prediction problem so that a new SemGCN  \cite{zhao2019semantic} is proposed in this field. However, GCN frameworks are never introduced for keypoints detection problem such as MPPE. In comparison, our graph network is specially designed for keypoints detection and contains a progressive learning strategy and guided by image features. 


\section{OPEC-Net: Occluded Pose Estimation and Correction}
\label{sec:network}
Existing pose estimation approaches achieve striking results on visible joints but produce wildly inaccurate outcomes on invisible ones. This is mainly due to the fact that localizing invisible joints from the heatmap is very challenging since they are occluded and there is a lack of visual information. To rectify this shortcoming, we introduce a novel framework that infers invisible joints from the image contextual and pose structural clues. 

Considering that, we generate an initial pose from a heatmap-based module and process it into an GCN-based joints correction module to learn their precise position. An Image-Guided GCN network (IGP-GCN) and a Cascaded Feature Adaption module is proposed in the correction stage. The IGP-GCN network exploits the human body structure and image context together to optimize the estimation results. By learning the displacements in a progressive way, it also offers a stable way to achieve more accurate results. 

\begin{figure*}[h]
\begin{center}
   \includegraphics[width=1\linewidth]{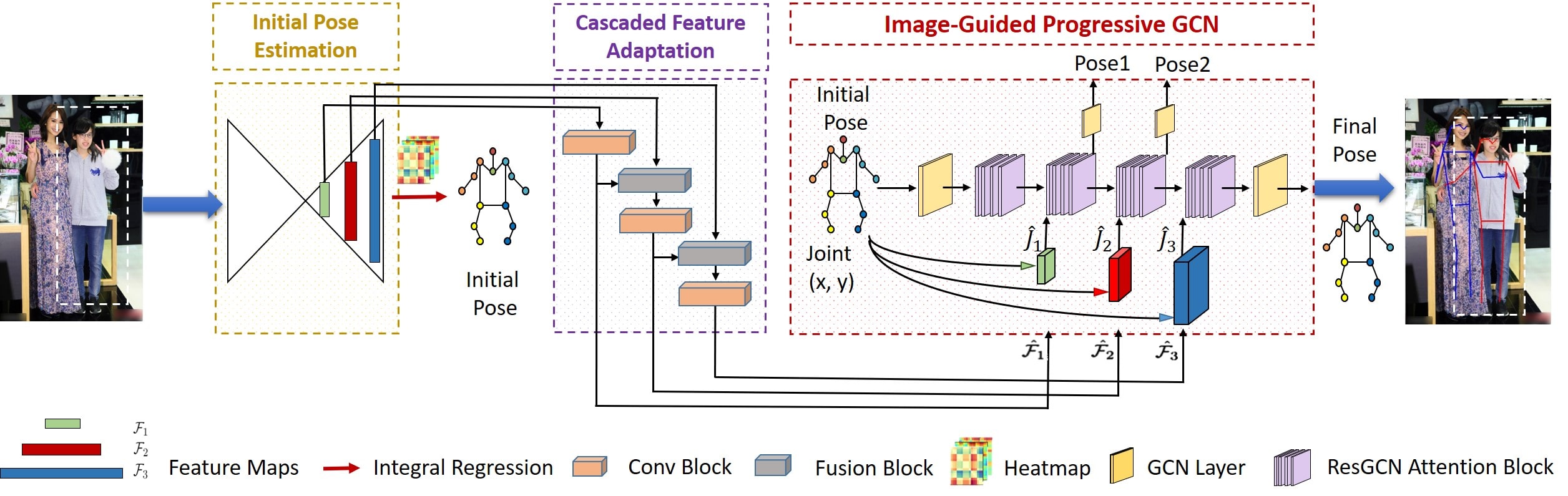}
\end{center}
   \caption{\textbf{The schematic diagram of our pipeline.} This figure depicts the two stages of estimation for one single pose. The GCN-based pose correction stage contains two modules: the Cascaded Feature Adaptation and the Image-Guided Progressive GCN. Firstly a base module is employed to generate heatmaps. After that, an integral regression method \cite{sun2018integral} is employed to transform the heatmap representation into a coordinate representation, which can be the initial pose for GCN network. The initial pose and the three feature maps from the base module are processed in Image-Guided Progressive GCN. The multi-scale feature maps are updated through the Cascaded Feature Adaptation module and put into each ResGCN Attention blocks. $\hat{J}_1$, $\hat{J}_2$ and $\hat{J}_3$ are the node features excavated on related location $(x, y)$ from image features. The error of Initial Pose, Pose1, Pose2, and Final Pose are all considered in the objective function. Then the OPEC-Net is trained entirely to estimate the human pose. The details of the whole framework are described in Section \ref{sec:network}}
\label{fig:pipeline}
\end{figure*}

The heatmap and coordinate modules in our framework are actually interdependent. Due to our heatmap inference network, the IGP-GCN module has a more accurate pose initialization, which also contributes to a more precise local contextual understanding, before conducting corrections. On the other hand, coordinate based IGP-GCN also addresses the limitation of heatmap modules: due to a size limit, heatmap representation usually causes quantisation error for joints estimations. Our IGP-GCN design tackles this issue by converting the heatmap into coordinate representation. The overall framework and the proposed OPEC-Net module is illustrated in Fig \ref{fig:pipeline}.

\subsection{Initial Pose Estimation from Heatmap-based modules}

In this stage, AlphaPose+ \cite{li2019crowdpose} is employed as the base module to generate a heatmap for visible joints. This is a top-down approach, which first detects a bounding box for each person and then performs instance-level human pose estimation. We describe the process for an instance-level human pose in the following.

Firstly, the three layers of the decoder of the base module generate three corresponding feature maps with different levels of fine details: a coarse feature map $\mathcal{F}_1$, a middle feature map $\mathcal{F}_2$ and a fine feature map $\mathcal{F}_3$. The base module outputs a heatmap which has high confidence for visible joints. The estimated pose from the heatmap $H$ can be denoted as $P$, which contains estimation results for each joint: 
\begin{equation}
    \{ <x^1, y^1, c^1>,\ <x^2, y^2, c^2>,\ ..., <x^k, y^k, c^k>\}    
\end{equation}
where $x^j$ and $y^j$ are the position of the $j$th joint, $c^j$ the confidence score, and $k$ is the number of joints in the skeleton. 

\subsection{GCN-based Joints Correction}
The occluded poses can be inferred easily by humans mostly because of their abundant prior knowledge of implicit body structure and pose properties. More specifically, a natural human pose is highly constrained by the environments and human body property, such as the biomechanical structure of the human body and implications in the environments. In light of that, we propose an Image-Guided graph network for correction which takes the initial pose generated from the above modules and adjusts the estimation results according to the implicit relationship of joints. 

\subsubsection{Heatmap representation to Coordinate representation.}
First of all, we generate the initial pose for the GCN network from the heatmaps of the first two stages. An important factor to consider in obtaining the initial pose is that the translation from heatmap to coordinate representation needs to be differential for the end to end training purpose, so the initial pose cannot be grasped directly from the heatmap by searching max values as $P$. Finally, we find out that a coordinate initial pose $\hat{J}_i$ can be generated from $Heatmap$ and estimated by an $integral\ regression$ method \cite{sun2018integral}.

Specifically, the heatmap is propagated into a Softmax layer which normalizes the values into likelihood values $[0,\ 1]$. After that, an integral operation is applied on the likelihood map to sum up the values and estimate joints positions. 

\begin{equation}
    \hat{J}_i^k = \int_{p \in A} p \cdot H_k(p),
\end{equation}

where $\hat{J}_i^k$ is the position estimation of the $k$th joint. We use $A$ to denote the region of likelihood and $H_k(p)$ to represent the likelihood value on point $p$. Therefore, every heatmap matrix contains the information to produce an initial pose $P_{init}$. 

\subsubsection{Graph Formulation.} The human body skeleton has a natural hierarchical graph structure. Previous researches on MPPE merely utilize this information by a primitive graph matching strategy. We claim that the implicit relationships between different joints are helpful to guide position estimation. We thus construct an intuitive graph $G = (V, E)$ to formulate the human pose with $N$ joints. $V$ is the node set in $G$ which can be denoted as $V = \{v_i\ | i = 1,\ 2,\ ...,\ N\}$. $E = \{v_iv_j\ |\ \text{if $i$ and $j$ are connected in the human body}\}$ is the edge set which refers to limbs of the human body. The adjacent matrix of $G$ refers to matrix $A = \{ a_{ij}\}$, with $a_{ij} = 1$ when $v_i$ and $v_j$ are neighbors in $G$ or $i=j$, otherwise $a_{ij} = 0$. 

For every node, the input feature $G_i^j$ is the joint estimation result $<x_i^j, y_i^j, c_i^j>$, where $i$ is $i$th pose and $j$ is the $j$th joint of the skeleton. We denote $G_i \in \mathbb{R}^{L \times N}$ as the input feature of the $i$th pose in the training set, where $L$ is the feature dimension.

\subsubsection{Image-Guided Progressive GCN Network.}
The core methodology proposed in our work is the Image-Guided Progressive GCN for Correction. In this network, the image context and pose structure clues for invisible joints inference are merged together in an innovative way. The details of each layers and ResGCN Attention Blocks are describe in supplementary materials.

(1) The estimated position of invisible joints from the base module is sometimes far from their correct locations and this makes it challenging to directly regress their displacements. Therefore, we design an intuitive coarse-to-fine learning mechanism in the coordinate-based module, which builds a \textbf{\textit{progressive GCN architecture}} and leverages the performance steadily by enforcing multi-scale image features in a progressive manner.

(2) The coordinate-based module lacks \textbf{\textit{local context information}}. Consequently, we excavate the related image features for each joints position and fuse them into the module. In another word, we improve the pose estimation results by incorporating image feature maps $\hat{\mathcal{F}_1}, \hat{\mathcal{F}_2}$, and $\hat{\mathcal{F}_3}$. Specifically, we design cascaded ResGCN attention blocks to grasp the useful information that is stored in the feature maps but lost in initial pose $\hat{P}_i$. The three feature maps are ordered from coarse to fine according to their size of receptive fields. After that, we employ a grid sample method that obtains the $j$th joint feature by excavating the feature located in $<x_i^j, y_i^j>$ on the related coordinate weight feature map. Every pose leads to three node feature vectors $\hat{J}_{1}$, $\hat{J}_{2}$, and $\hat{J}_{3}$ extracted following this process. Finally, these node features are fed into the ResGCN attention blocks accordingly.


\subsubsection{Cascaded Feature Adaption (CFA).}
Feature maps $\mathcal{F}_1,\mathcal{F}_2$, and $\mathcal{F}_3$ should be adaptive to provide more effective information to the IGP-GCN. Moreover, the low-level feature and high-level feature are fused in the cascaded design in order to enlarge their respective fields resulting the updated feature are more informative. The details of Conv Blocks and Fusion Blocks used in this module is in supplementary materials.

\subsubsection{CoupleGraph}
\label{CoupleGraph}
We extend the single human graph into a CoupleGraph that captures more human interactions and this is achieved by connecting the corresponding joints to capture human interaction information. The couple graph can be denoted as $G' = (V', E')$. The joints number of a single person is $N$ so that there is $2N$ joints in total in the couple graph. It can be formulated as $V' = \{v_i\ | i = 1,\ 2,\ ...,\ 2N\}$. There are two types of edges in $E'$, the edges representing the human skeleton $E_s$ and the edges connecting the two humans $E_c$. The human skeleton edges are noted as $E_s = \{v_iv_j\ |\ \text{if $i$ and $j$ are connected in the human body}\}$. The human interaction edges can be written as $E_c = \{v_iv_{i+N}\}$, where the $v_i$ and $v_{i+N}$ are correspond to the same components of the two human skeletons. The CoupleGraph module is appended after OPEC-Graph module to enhance the performance of estimation. Each pair of people is processed by CoupleGraph.

\subsection{Loss Functions}
\label{sec:loss}
The objective function of our OPEC-Net module can now be formulated.  We denote the training set as $\Omega$, the ground truth pose in $\Omega$ as $P_i$, and the output pose of $j$th ResGCN Attention block as $\hat{P}_{ij}$. From heatmap representation to coordinate representation, the integral regression method produces an initial pose $\hat{P}_{init}$. Hence, the total loss is defined as the sum of the rectified loss of poses from IGP-GCN and initial loss of initial pose:

\begin{equation}
    Loss = \mathop{min}\limits_{\theta} \sum_{i \in \Omega}(\sum_{j=1}^n \lambda_j | (\hat{P}_{ij} -P_i)  \odot M | + |(\hat{P}_{init} - P_i) \odot M|)
\end{equation}

The term $| \hat{P}_{ij} -P_i |$ indicates the calculation of the $L_1$ loss between our estimated pose and the ground truth, $n$ is the number of ResGCN attention blocks in the model. In this work, we set $n = 3$. We sum up all the errors of the produced pose from each block and assign a parameter $\lambda_j$ to control the weight. All the trainable parameters in our network are denoted as $\theta$. $M \in \mathbb{Z}_2^{N}$ is a binary mask where the element in $M$ corresponds to 1 when the related joint has a ground truth label, otherwise it is 0. The $\odot$ denotes the element-wise product operation so that we only take into account the errors on the joints with ground truth.

The lasted generated pose will correspond to the best estimation result, so we treat the final one as our estimated result. 

\section{Our Occluded Pose Dataset}

We build a new dataset, called \textbf{Occluded Pose} (OCPose), that includes more heavy occlusions to evaluate the MPPE. It contains challenging invisible joints and complex intertwined human poses. We mostly consider the couple pose scenes, such as dancing, skating, and wrestling, because they have more reliable annotations and practical utility. This section gives details of data collection, data annotation, and data statistics.

\textbf{Data Collection.}
The ground truth of human pose can be hard to recognize when the occlusions are extremely heavy. Thus, we majorly collect videos of two-person interactions since they are much easier to annotate because the volunteer can infer the pose from contextual information. We first search for videos from the Internet by using keywords such as boxing, dancing, and wrestling. We then capture the distinctive images which contain diverse poses and humans from these videos by restricting the interval to be at least 3 seconds. Finally, we manually sift through the clips to select high-quality images. All the images are collected under the permission of privacy issues. 

\textbf{Data Annotation.}
We develop an annotation tool for the user to bound the area of the couple and then locate two template skeletons to their right positions. Six volunteers are recruited for manual labelling. Each skeleton has 12 joints and the left and right components are distinct. In addition to annotating the bounding box and the human body poses, the volunteers also need to indicate whether the joint is visible or not. To ensure accuracy, we use cross annotation for every image. At least two volunteers are required to provide their annotations on the same image. If an intolerant deviation exists between their results, the image is annotated again. The final joint positions are the mean value of the two annotations.

\begin{table}[h]
\begin{center}
\caption{The comparison of occlusion level. We count the number of images of each dataset with different level of occlusion. As shown above, MSCOCO and MPII almost have no heavily occlusions. OCHuman is the state-of-the-art dataset for occlusions but our dataset is larger and contains more severe occlusion}
\label{tab1:IoU}
\resizebox{0.8\textwidth}{!}{
\begin{tabular}{llllll}
\Xhline{1.2pt}
\noalign{\smallskip}
Dataset & Total & IoU$>$0.3 & IoU$>$0.5 & IoU$>$0.75 & Average\\
\noalign{\smallskip}
\hline\hline
\noalign{\smallskip}
\textbf{CrowdPose} & 20000 & 8706 (44\%) & 2909 (15\%) & 309 (2\%)  & 0.27  \\
\textbf{MSCOCO} & 118287 & 6504 (5\%) & 1209 (1\%) & 106 ($<$1\%) & 0.06 \\
\textbf{MPII} & 24987 & 0 & 0 & 0 & 0.11 \\
\textbf{OCHuman} & 4773 & 3264(68 \%) & 3244(68\%) & 1082(23\%) & 0.46 \\
\noalign{\smallskip}
\hline
\noalign{\smallskip}
Ours & 9000 & 8105 \textbf{(90\%)} & 6843 \textbf{(76\%)} & 2442 \textbf{(27\%)} & \textbf{0.47}\\
\noalign{\smallskip}
\Xhline{1.2pt}
\end{tabular}
}
\end{center}
\end{table}

\textbf{Data Statistics.}
In total, our dataset contains 9000 images and 18000 fully annotated persons. For the training process, the training dataset consists of 5000 images, whereas validation and test dataset each contains 2000 images. 

To compare the occlusion level, we evaluate the average intersection over union (IoU) of bounding box on the other public benchmarks, such as CrowdPose \cite{li2019crowdpose}, OCHuman \cite{zhang2019pose2seg}, MSCOCO \cite{lin2014microsoft} and MPII\cite{andriluka20142d}. We report the comparison result of these benchmarks in Table \ref{tab1:IoU}, which illustrates that our dataset beats down all the other benchmarks on the occlusion level.

\textbf{Other Dataset.} In our approach, we carried out extensive experiments on public benchmarks. Following the typical training procedure, we evaluate the OPEC-Net on our OCPose, CrowdPose \cite{li2019crowdpose}, MSCOCO \cite{lin2014microsoft} and particular occluded dataset OCHuman \cite{zhang2019pose2seg}. CrowdPose dataset is split in a ratio of $5:4:1$ for training, testing and validation respectively. We regard the validation set of OCHuman with 2500 images as our training dataset, and the rest 2273 images for testing. Then we follow the typical training strategy on MSCOCO. 

\section{Experiments}
In this section, extensive quantitative and qualitative experiments are demonstrated to evaluate the effectiveness of our OPEC-Net. Comprehensive ablation studies are carried out to validate the effectiveness of each components. 


\subsection{Experiments Settings}

\textbf{Implementation Details.} 
For training, we set the parameters $\lambda_1 =0.3,\ \lambda_2 =0.5,\ \lambda_3 = 1$ and $epochs = 30$. We feed $10$ images in a batch to train the whole framework. The initial learning rate is set to $1e^{-3}$ and decays in a cosine way. The input image size are $384 \times 288$ for MSCOCO and $320 \times 256$ for the other datasets. An AdamOptimizer is employed to optimize the parameters by backpropagation. For a fair comparison, we filter the proposal of the instances in the background and only focus on the Object Keypoint Similarity (OKS) of targets when we evaluate baselines on our dataset. We implement our model in PyTorch \cite{paszke2017automatic} and conduct experiments on one Nvidia GeForce GTX 1080 Ti with 11GB memory. More details are described in the supplementary materials. 

\textbf{Evaluation Metric.}
We follow the standard evaluation metric of MSCOCO, which is widely used by existing work as well \cite{fang2017rmpe,li2019crowdpose,huang2019mask,cao2017realtime}. Specifically, we report the mean Average Precision (mAP) value at 0.5:0.95, 0.5, 0.75, 0.80 and 0.90. In order to grasp the qualified poses for OPEC-Net training procedure, two rules are formulated to select the proposal. The proposal poses must contain more than 5 visible points and OKS value more than $0.3$ to ensure the quality. To enrich the dataset, we also flip the images as a data augmentation strategy. Furthermore, we provide the visualization results of pose estimation.

\textbf{Baselines.}
For comparison, we assess the performance with our OPEC-Net module using the three state-of-the-art approaches for MPPE: Mask RCNN \cite{he2017mask},
AlphaPose+\footnote{\url{https://github.com/MVIG-SJTU/AlphaPose/tree/pytorch.}} \cite{li2019crowdpose} and SimplePose \cite{xiao2018simple}. For a fair comparison, we quote the results of Mask RCNN and SimplePose directly from paper \cite{li2019crowdpose} and re-train AlphaPose+ from their public code. For the evaluation on OCPose, CrowdPose and OCHuman, we take AlphaPose+ for the initial pose estimation stage with ResNet-101 as backbone and Yolo V3 as detector. For MSCOCO dataset, we take the public code of SimplePose\footnote{\url{https://github.com/leoxiaobin/pose.pytorch.}} for the first stage for it has higher performance than AlphaPose+ on MSCOCO. Mask RCNN is used as the detector and ResNet-152 is used as the backbone on MSCOCO. OPEC-Net here denotes the framework with a single person as graph, and CoupleGraph denotes the baseline that performs a CoupleGraph based framework after OPEC-Net. 

\begin{figure*}[h]
\begin{center}
   \includegraphics[width=1\linewidth]{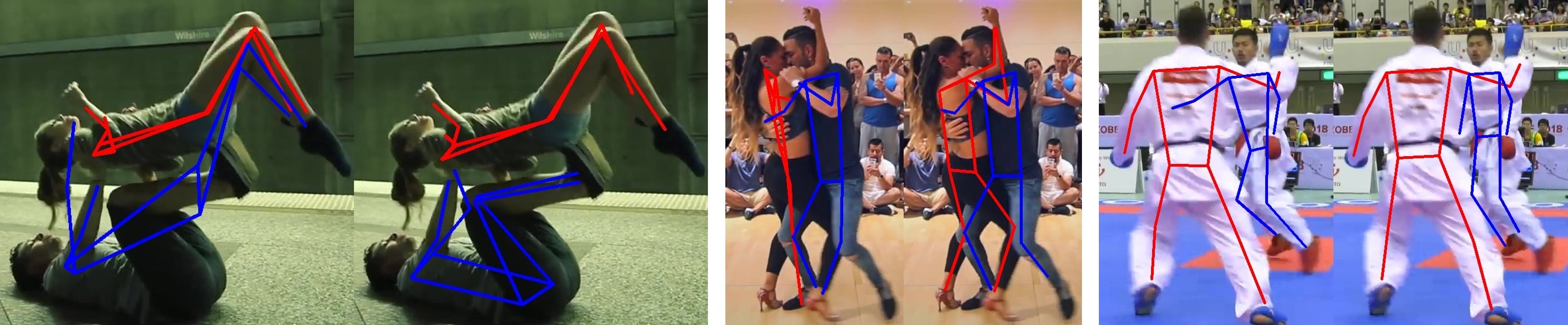}
\end{center}
   \caption{The qualitative evaluation of CoupleGraph and OPEC-Net. The left images are generated from OPEC-Net and the right ones come from CoupleGraph }
\label{fig:CG}
\end{figure*}

\begin{table}[htpb]
    \caption{The comparison on our OCPose dataset}
    \label{tab:OurmAP}
    \centering
    \resizebox{0.9\textwidth}{!}{
    \begin{tabular}{p{5.5cm}<{\centering}p{3cm}<{\centering}p{1.5cm}<{\centering}p{1.5cm}<{\centering}p{1.5cm}<{\centering}p{1.5cm}<{\centering}}
    \Xhline{1.2pt}
    \noalign{\smallskip}
    Method & mAP@0.5:0.95 & $AP^{50}$ & $AP^{75}$ & $AP^{80}$ & $AP^{90}$\\
    \noalign{\smallskip}
    \hline
    \hline
    \noalign{\smallskip}
    \textbf{Mask RCNN} \cite{he2017mask} & 21.5 & 49.8 & 15.9 & 7.7 & 0.1\\
    \textbf{Simple Pose} \cite{xiao2018simple} & 27.1 & 54.3 & 24.2 & 16.8 & 4.7\\
    \textbf{AlphaPose+} \cite{li2019crowdpose} & 30.8 & 58.4 & 28.5 & 22.4 & 8.2\\
    \hline
    \textbf{OPEC-Net} & 32.8(+2.0) & 60.5 & 31.1 & 24.0 & 9.2\\
    \textbf{CoupleGraph} & \textbf{33.6(+2.8)} & \textbf{60.8} & \textbf{32.5} & \textbf{25.0} & \textbf{9.8} \\
    \noalign{\smallskip}
    \Xhline{1.2pt}
    \end{tabular}
    }
\end{table}

\begin{figure*}[t]
\begin{center}
   \includegraphics[width=1\linewidth]{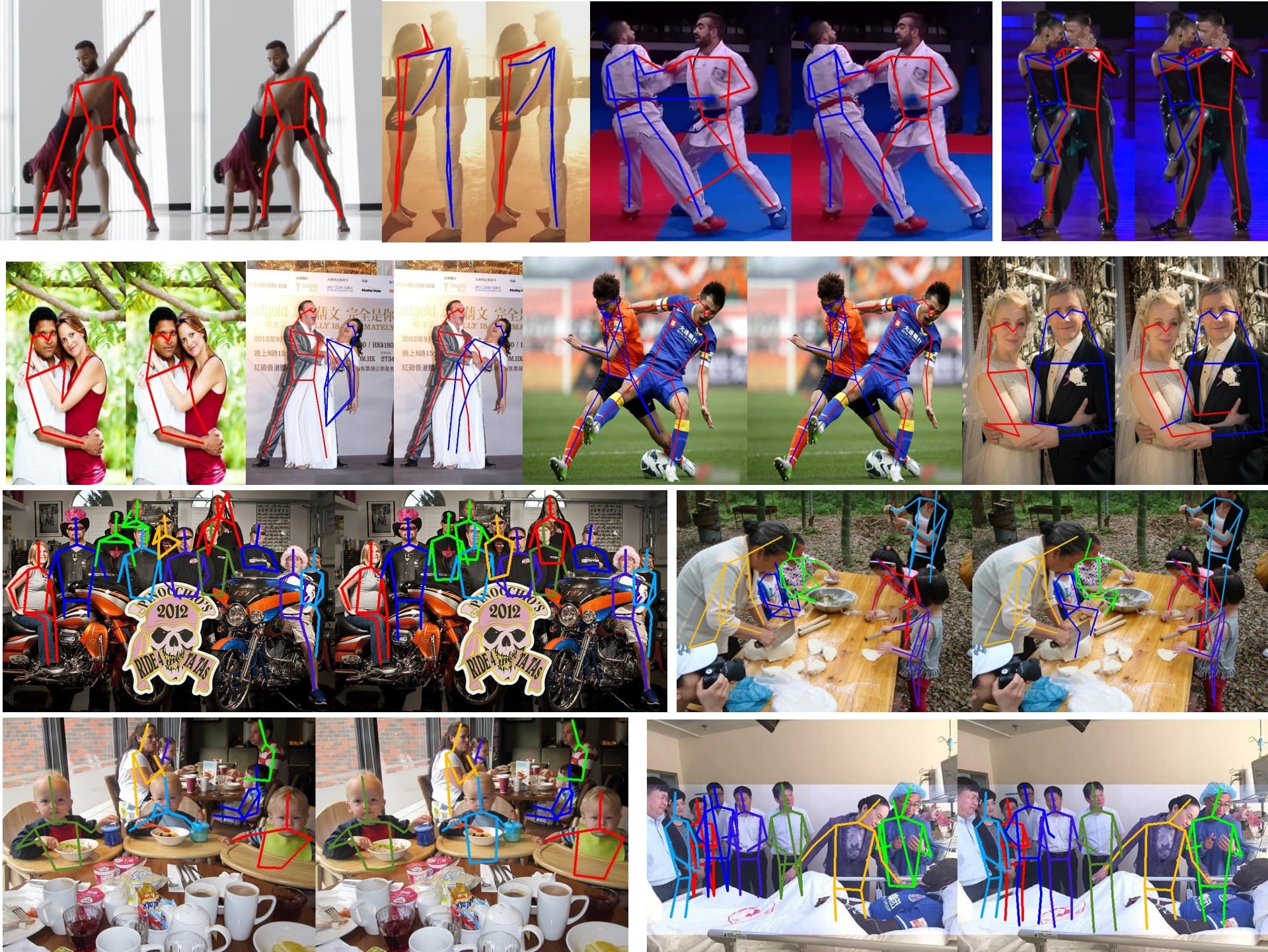}
\end{center}
 \caption{The results on OCPose, OCHuman and CrowdPose.  These are the qualitative comparison results of AlphaPose+ method and OPEC-Net on our datasets. The left pose is estimated by AlphaPose+ method and the right one is ours. The first row is OCPose and the second row represents OCHuman, the rest represents CrowdPose}
\label{fig:threeresults}
\end{figure*}

\subsection{Performance Comparison on our OCPose dataset}
\textbf{Quantitative Comparison.} The quantitative results are presented in Table \ref{tab:OurmAP}. Our approach attains the best mAP comparing to all the baselines with a considerable margin. OPEC-Net achieves a significant gain which is surprisingly 2.0 mAP@0.5:0.95 improvement compared to AlphaPose+. Despite of that, a significant 1.0 $AP^{90}$ improvement has been achieved which proves that our OPEC-Net has great ability of inference especially for high level of occlusions compared to localization methods. In conclusion, these results validate the prominent effectiveness of our OPEC-Net module on MPPE tasks. 

\textbf{Qualitative Comparison.} As illustrated in the first row of Fig \ref{fig:threeresults}, our OPEC-Net is capable of correcting the wrong link between joints and estimating the occluded joints while maintaining high performance on visible joints. We make these observations from the results: (1) For the first sample, a superior pose estimation result is provided by our method. Even an error with large displacement can be corrected by OPEC-Net. (2) Moreover, although the second case has difficult sunlight interference, our approach can also adjust the joints to their correct locations. (3) The third group also shows an evidence that our OPEC-Net module produces more natural poses that conform to human body constraints. (4) The fourth figure shows that our method can find the correct link between joints. 

\textbf{CoupleGraph.} The evaluations of CoupleGraph are given in Tab \ref{tab:OurmAP} and Fig \ref{fig:CG}. Comparing to OPEC-Net, CoupleGraph baseline also shows an advanced lifting 0.8 mAP@0.5:0.95, which validates the human interaction clues are quite prominent. As illustrated in Fig \ref{fig:CG}, CoupleGraph outperforms OPEC-Net significantly in quality. In these human interactive scenarios, the poses estimated by CoupleGraph are more concordant and superior. 

\subsection{Comparison against state-of-the-Arts on other benchmarks}
Extensive evaluations on heavily benchmarked dataset demonstrate the effectiveness of our model for occlusion problem. The experimental results on existing benchmarks are presented in Table \ref{tab:TwomAP}, \ref{tab:COCOmAP} and Fig \ref{fig:threeresults}. Our model surpasses all the baselines by a considerable margin.

\begin{table*}[h]
\renewcommand\arraystretch{1.12}
\caption{The qualitative result on occlusion dataset}
\label{tab:TwomAP}
\begin{center}
\resizebox{0.95\textwidth}{!}{
\begin{tabular}{l|c|c|c|c|c!{\vrule width1.2pt}c|c|c|c|c}
\Xhline{1.2pt}
\multirow{2}*{\diagbox{Method}{Dataset}} &\multicolumn{5}{c!{\vrule width1.2pt}}{OCHuman \cite{zhang2019pose2seg}}& \multicolumn{5}{c}{CrowdPose \cite{li2019crowdpose}}  \\
\cline{2-11}
& mAP@0.5:0.95 & $AP^{50}$ & $AP^{75}$ & $AP^{80}$ & $AP^{90}$ & mAP@0.5:0.95 & $AP^{50}$ & $AP^{75}$ & $AP^{80}$ & $AP^{90}$ \\
\hline
\hline
\textbf{Mask RCNN} \cite{he2017mask} & 20.2 & 33.2 & 24.5 & 18.3 & 2.1 & 57.2 & 83.5 & 60.3& - & -  \\
\hline
\textbf{SimplePose} \cite{xiao2018simple} & 24.1 & 37.4 & 26.8 & 22.6 & 4.5 & 60.8 & 81.4 & 65.7& - & -  \\
\hline
\textbf{AlphaPose+} \cite{li2019crowdpose} & 27.5 & 40.8 & 29.9 & 24.8 & 9.5 & 68.5 & 86.7 & 73.2& 66.9 & 45.9  \\
\hline
\textbf{OPEC-Net} & \textbf{29.1(+1.6)} & \textbf{41.3} & \textbf{31.4} & \textbf{27.0} & \textbf{12.8(+3.3)} & \textbf{70.6(+2.1)} & \textbf{86.8} & \textbf{75.6}& \textbf{70.1} & \textbf{48.8(+2.9)}  \\
\hline
\Xhline{1.2pt}
\end{tabular}
}
\end{center}
\end{table*}

\noindent \textbf{OCHuman} \hspace{0.3cm} As OCHuman is a new benchmark proposed mainly for pose segmentation, we are the first to report all the baseline results on this challenging occlusion dataset. Comparing to AlphaPose+, we achieve maximal 3.3 improvements on $AP^{90}$. This further validates that our OPEC-Net model is robust even for highly challenging occlusion scenarios.  

\noindent \textbf{CrowdPose} \hspace{0.3cm} As shown in Table \ref{tab:TwomAP}, OPEC-Net drastically lifts 2.1 mAP@0.5:0.95 of the estimation result over AlphaPose+. It is also worth noting that the improvements remain high when the comparison AP terms are high. For example, our model achieves 0.1, 2.4, 3.2 and 2.9 on AP 50, 75, 80 and 90 respectively.

\noindent \textbf{MSCOCO} \hspace{0.3cm} We also present our results on the largest benchmark MSCOCO. Our model only contributes slightly accuracy improvements. The reason is the key difference between MSCOCO and other datasets -- it contains too few occlusion scenarios, especially the severe ones. Moreover, a lot of invisible joints lack annotations on MSCOCO. 
\begin{table}
    \caption{\textbf{MSCOCO} 2017 test-dev set
    \cite{lin2014microsoft}}
    \label{tab:COCOmAP}
    \centering
    \resizebox{0.8\textwidth}{!}{
    \begin{tabular}{p{3cm}<{}p{3cm}<{\centering}p{3cm}<{\centering}p{3cm}<{\centering}}
    \Xhline{1.2pt}
    \noalign{\smallskip}
    Method  & mAP@0.5:0.95 & mAP@0.5 & mAP@0.75 \\
    \noalign{\smallskip}
    \hline
    \hline
    \noalign{\smallskip}
    \textbf{AlphaPose+} \cite{li2019crowdpose} & 72.2 & 90.1 & 79.3\\
    \hline
    \textbf{Simple Pose} \cite{xiao2018simple} & 73.7 & 91.9 & 81.8\\
    \hline
    \textbf{OPEC-Net} & \textbf{73.9(+0.2)} & \textbf{91.9} & \textbf{82.2} \\
    \Xhline{1.2pt}
    \end{tabular}
    }

\end{table}

\noindent \textbf{Invisible \textit{vs.} Visible} \hspace{0.3cm} To investigate of the effectiveness on invisible (Inv) and visible (V) joints separately, we report the statistics of each type of joints according to the similar rule of OKS. From Tab \ref{tab:twojoints}, our OPEC-Net improves mostly on the invisible joints rather than visible joints. In terms of Inv@75, our framwork achieves a considerable marginal 3.3\% and 4.9\% gains on CrowdPose and OCPose respectively. On the contrary, the OPEC-Net only improves a maximal 1\% on visible joints because our main focus is the invisible joints. This comparison also explains why the gains are smaller on MSCOCO datasets than the other datasets that contains more occlusions. 

\begin{table}[h]
    \caption{Results for Visible and Invisible Joints on CrowdPose and OCPose}
    \label{tab:twojoints}
    \centering
    \resizebox{1\textwidth}{!}{
    \begin{tabular}{p{1.7cm}<{}|p{1.7cm}<{\centering}|p{1.7cm}<{\centering}!{\vrule width1.2pt}p{1.7cm}<{\centering}|p{1.7cm}<{\centering}!{\vrule width1.2pt}p{1.7cm}<{\centering}|p{1.7cm}<{\centering}!{\vrule width1.2pt}p{1.7cm}<{\centering}|p{1.7cm}<{\centering}}
    \Xhline{1.2pt}
    Datasets &\multicolumn{4}{c!{\vrule width1.2pt}}{CrowdPose} & \multicolumn{4}{c}{OCPose}\\
    \Xhline{1.2pt}
    Method & Inv@75 & Inv@90 & V@75 & V@90 & Inv@75 & Inv@90 & V@75 & V@90\\
    \hline
    \hline
    AlphaPose+ & 76.2\% & 57.2\% & 89.5\% & 67.8\% &50.7\% & 17.7\% & 85.2\% & \textbf{55.3}\%\\
    \hline
    OPEC-Net & \textbf{79.5}\% & \textbf{58.4}\% & \textbf{90.0}\% & \textbf{67.8}\% &\textbf{55.6}\% & \textbf{20.5}\% & \textbf{86.2}\% & 55.1\%\\
    \Xhline{1.2pt}
    \end{tabular}
    }
\end{table}

\subsection{Alabtion studies}
To analyze our model in details, we conduct comprehensive ablative experiments to evaluate the capability of each component and clues we claimed. As illustrated in Table \ref{tab:Ablation}, we present the baselines to investigate the impact of each component. 

Firstly, we investigate the impact of image guided strategy that blends the image context with GCN. From (a), a clear decrease around 2.0 mAP@0.5:0.95 is observed, which points out the importance of the Image-Guide strategy. Without the Image-Guided part, a single GCN network improves the performance poorly. This evidence validates that the GCN module must learn under the guidance of image features. 

We further investigate each design of the IGP-GCN. From (b) and (c), we can conclude that the strategy of progressive and coarse to fine feature learning is effective. Moreover, the proposed Cascaded Feature Adaption module is analysed as well. In Table \ref{tab:Ablation}, the mAP value of three datasets falls down significantly, demonstrating that the CFA module plays an indispensable role in the whole framework. We remove the Fusion Blocks and report the results in (c), which further proves the effectiveness of the fusion part in the CFA module.   
We can conclude that the image guidance is the most imperative in the framework. The CFA module brings an average 0.7 mAP@0.5:0.95 gain on three datasets, which manifests the necessity to make image features adaptive for coordinate module. 
Overall, these ablation studies overwhelmingly validate that every component is effective and the clues are informative for invisible joints inference.

\begin{table}[h]
    \caption{Ablation study of our OPEC-Net framework (mAP@0.5:0.95)}
    \label{tab:Ablation}
    \centering
    \resizebox{0.95\textwidth}{!}{
    \begin{tabular}{p{7.5cm}<{}p{1.5cm}<{\centering}p{1.5cm}<{\centering}p{1.5cm}<{\centering}}
    \Xhline{1.2pt}
    \noalign{\smallskip}
    The evaluation of removal each component & OCHuman & CrowdPose & OCPose \\
    \noalign{\smallskip}
    \hline
    \hline
    \noalign{\smallskip}
     The AlphaPose+ baseline & 30.8 & 27.5 & 68.5 \\
     (a)Without the Image-Guided strategy in GCN & 30.8 & 27.7 & 68.6 \\
     (b)Without the Progressive design in the GCN & 32.2 & 28.3 & 69.3 \\
     (c)Use one feature $F_3$ instead of multi-scale features  & 32.5 & 28.5 & 69.7 \\
     (d)Remove the Cascaded Feature Adaption module & 32.1 & 28.4 & 69.9 \\
     (e)Remove the Fusion Block in CFA module & 32.4 & 28.7 & 69.6 \\
     The full OPEC-Net & 32.8 & 29.1 & 70.6 \\
    \noalign{\smallskip}
    \Xhline{1.2pt}
    \end{tabular}
    }
\end{table}

\section{Conclusion}
In this paper, we proposed a novel OPEC-Net module and a challenging Occluded Pose (OCPose) dataset to address the occlusion problem in Crowd Pose Estimation. Two elaborate components, Image-Guided Progressive GCN and Cascaded Feature Adaptation, are designed to exploit the natural human body constraints and image context. We conduct thorough experiments on four benchmarks and ablation studies to demonstrate the effectiveness and provide a variety of insights. The heatmap and coordinate module are proved to work cooperatively and achieve remarkable improvements in all aspects. By making this dataset available, we hope to arouse the attention and increase the interest in the investigation of the occlusion problem in pose estimation.  

\section*{Acknowledgment}
\noindent The work was supported in part by grants No. 2018YFB1800800, No. 2018B030338001, No. 2017ZT0 7X152, No. ZDSYS201707251409055 and in part by National Natural Science Foundation of China (Grant No.: 61902334 and 61629101). The authors also would like to thank Running Gu and Yuheng Qiu for their early efforts on data labeling. 


%
%
\bibliographystyle{splncs04}
\bibliography{egbib}

~\\
~\\
~\\
~\\
~\\
~\\
\centerline{\LARGE{\emph{\textbf{Supplementary Materials}}}}
~\\
~\\
We provide details of our algorithms and more visualization results in this supplementary material. Two videos are attached for demonstrating the visual comparisons between our OPEC-Net and AlphaPose+ \cite{li2019crowdpose} in crowd and couple scenarios respectively. Our estimation displays as green skeletons and the red ones are from AlphaPose+. It is obvious to observe from the videos that our estimation results are more accurate than AlphaPose+, especially for occlusion cases. For the couple scenario, the comparison between OPEC-Net and OPEC-CG (with CoupleGraph extension) is also shown, where OPEC-Net is with red color while OPEC-CG is in green. As seen, with the help of CoupleGraph, the approach is more robust to severe occlusions. 

\setcounter{section}{0}
\section{Illustration of the CoupleGraph}
CoupleGraph is proposed to capture the human interaction information. In the main paper, we gave out the formulation of the CoupleGraph. Here, we illustrate the structure of the couple graph in the following:

\begin{figure}[h]
	\begin{center}
		\includegraphics[width=0.6\linewidth]{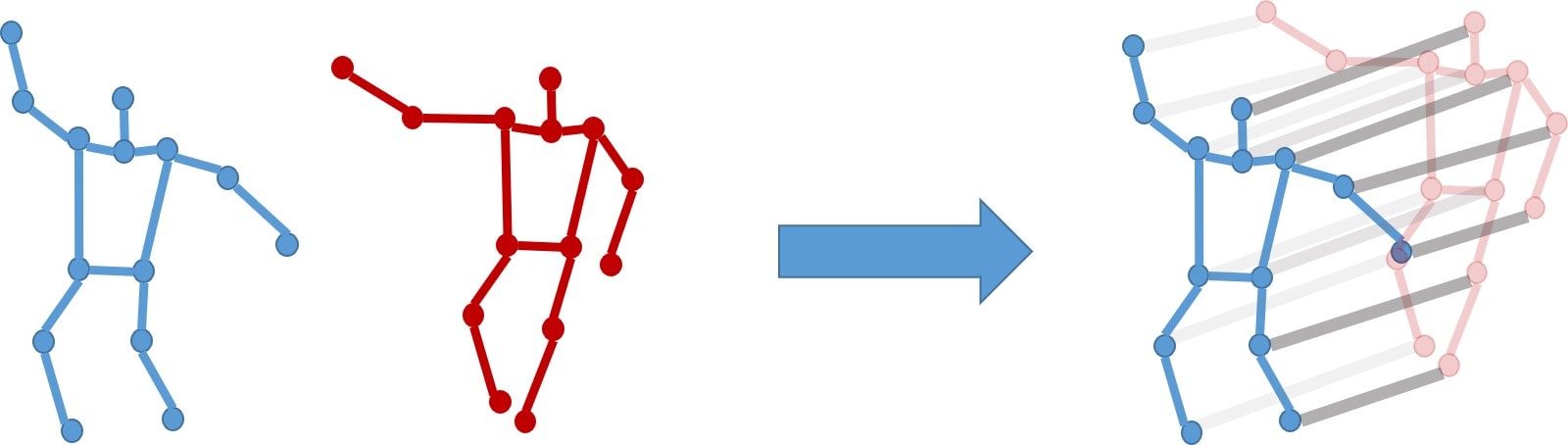}
	\end{center}
	\caption{The couple graph. The blue and red skeletons represent two different individuals. Other than in-skeleton bone edges, the graph also links the two corresponding joints between the two instances. }
	\label{fig:CoupleGraph}
\end{figure}


\section{Details of the Network Architecture}
We explain the detailed settings of the layers and parameters of our OPEC-Net in this section. Firstly, the proposals $J_k$ are normalized into the range of $[-1,\ 1]$. There is a ReLu layer after each deep GCN layer \cite{li2019deepgcns} in our network. 
In the module of Cascaded Feature Adaption (CFA), each convolutional layer is of a $3 \times 3$ kernel and is followed by a ReLu function. In IGP-GCN, the first and second GCN layers are of 128 channels. The first ResGCN Attention Block (RAB) takes a 128-channel feature map as input and outputs a feature map with 256 channels. Both the input and output channels of the following two RABs are in 256. The last block takes a 256-channel feature map together with a 128-channel feature map as input and outputs a 256-channel feature map. 

\section{ResGCN Attention Block}
\begin{figure}[h]
	\begin{center}
		\includegraphics[width=0.7\linewidth]{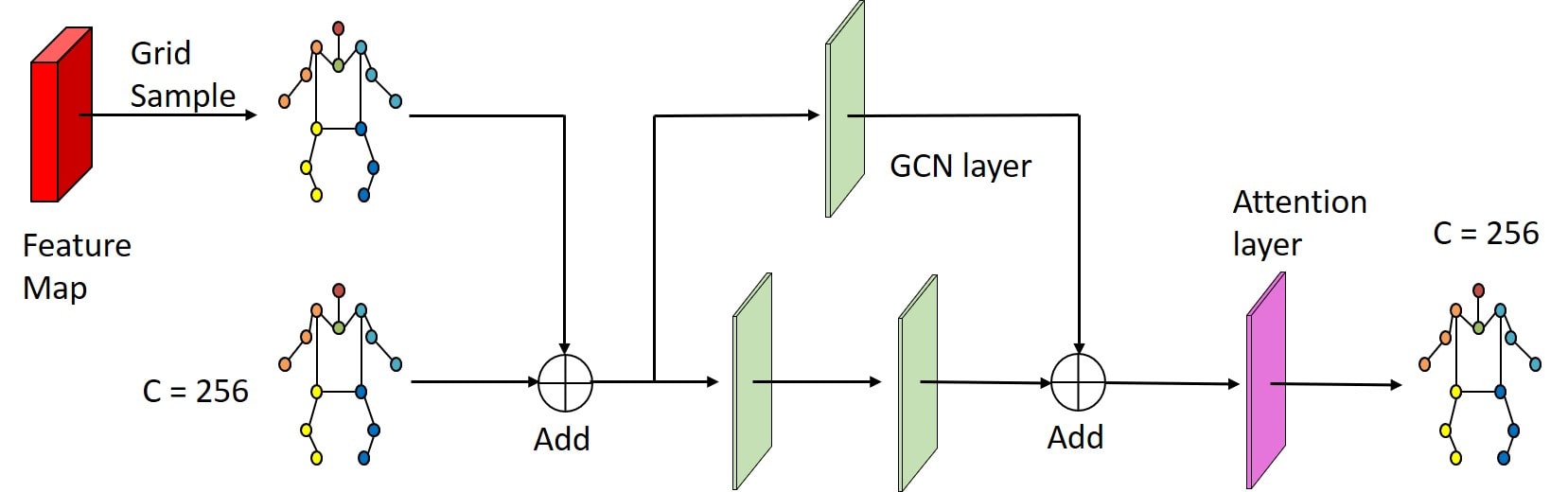}
	\end{center}
	\caption{The ResGCN Attention Block. $C$ denotes the channel size of the feature. 
	}
	\label{fig:ResGCN}
\end{figure}

As seen in Fig ~\ref{fig:ResGCN}, each RAB consists of three GCN layers\cite{li2019deepgcns} and a self-attention layer. To ease the learning, we also add a residual link in the RAB to make the network being focusing on inferring residual value. 
For this goal, before conducting addition with the output of the lower two GCN layers, we also involve a GCN layer (the upper one in the Figure) to adjust the feature size of the input. Furthermore, we append a self-attention layer at the end to learn the dependency of joints and utilize this to capture more informative understanding of the inherent relationships in the human pose, for more accurate estimation. 

\section{Cascaded Feature Adaption (CFA) Module}
\label{sec:PIN}

\begin{figure}
	\begin{center}
		\includegraphics[width=0.9\linewidth]{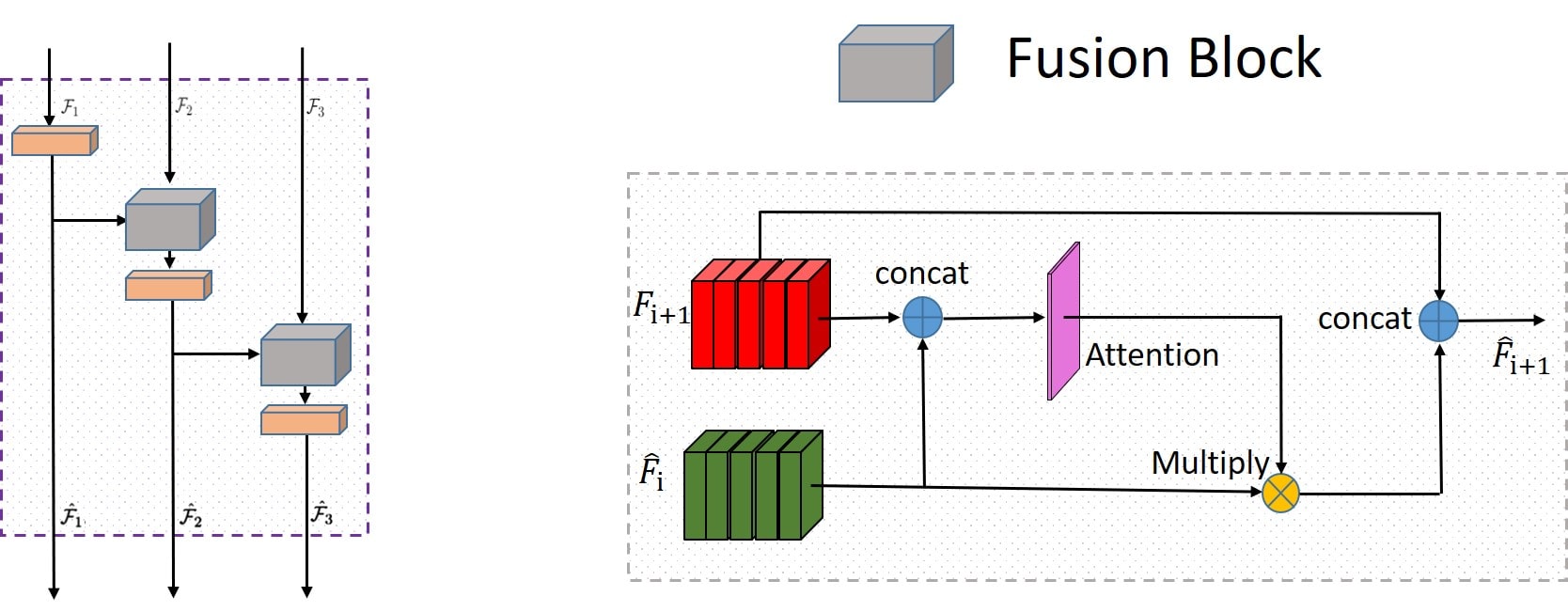}
	\end{center}
	\caption{The left figure shows the design of the Cascaded Feature Adaption module and the right one demonstrates the design of Fusion block. }
	\label{fig:CFA}
\end{figure}

As seen in Fig \ref{fig:CFA}, the CFA module consists of two Fusion blocks (Fig \ref{fig:CFA}) and three Conv blocks. In the experiments, we use one convolution layer in each Conv block. 
More specifically, the CFA module takes the three feature maps $\mathcal{F}_1,\mathcal{F}_2$, and $\mathcal{F}_3$ as input and outputs $\hat{\mathcal{F}_1}, \hat{\mathcal{F}_2}$, and $\hat{\mathcal{F}_3}$. Firstly, $\mathcal{F}_1$ is transformed to $\hat{\mathcal{F}_1}$ by a convolutional block. It can be formulated as

\begin{equation}
\hat{\mathcal{F}_1} = Conv(\mathcal{F}_1; \theta).
\end{equation}

We subsequently fuse two feature maps $\hat{\mathcal{F}_1}$ and $\mathcal{F}_2$ by a Fusion block, which gets a new feature map $\hat{\mathcal{F}_2}$ produced. This process will also be performed repeatedly again which generates $\hat{\mathcal{F}_3}$. We formulate this process as

\begin{equation}
\hat{\mathcal{F}}_{i+1} = Fusion(\hat{\mathcal{F}}_i, \mathcal{F}_{i+1}; \theta).
\end{equation}

Thanks to such CFA module, the feature maps $\hat{\mathcal{F}_1}, \hat{\mathcal{F}_2}$, and $\hat{\mathcal{F}_3}$ are more informative and adaptive, than  $\mathcal{F}_1,\mathcal{F}_2$, and $\mathcal{F}_3$, for the next stage. 

\begin{equation}
Heatmap = Conv(\hat{\mathcal{F}_3}; \theta).
\end{equation}

In particular, the Fusion Block takes $\hat{\mathcal{F}_i}$ and  $\mathcal{F}_{i+1}$ as input and outputs $\hat{\mathcal{F}}_{i+1}$. The two feature maps are concatenated and are passed into an attention layer, which learns a weighting way for automatically combining information from both sides. 


\section{Result Gallery}
\label{Result}
In this section, we present more visually comparison results on the three datasets: OCPose, OCHuman \cite{zhang2019pose2seg}, and CrowPose \cite{li2019crowdpose}. For each example, the left result is obtained from the AlphaPose+, and the right ones are estimated by our OPEC-Net.

\begin{figure*}[!h]
	\begin{center}
		\includegraphics[width=0.95\linewidth]{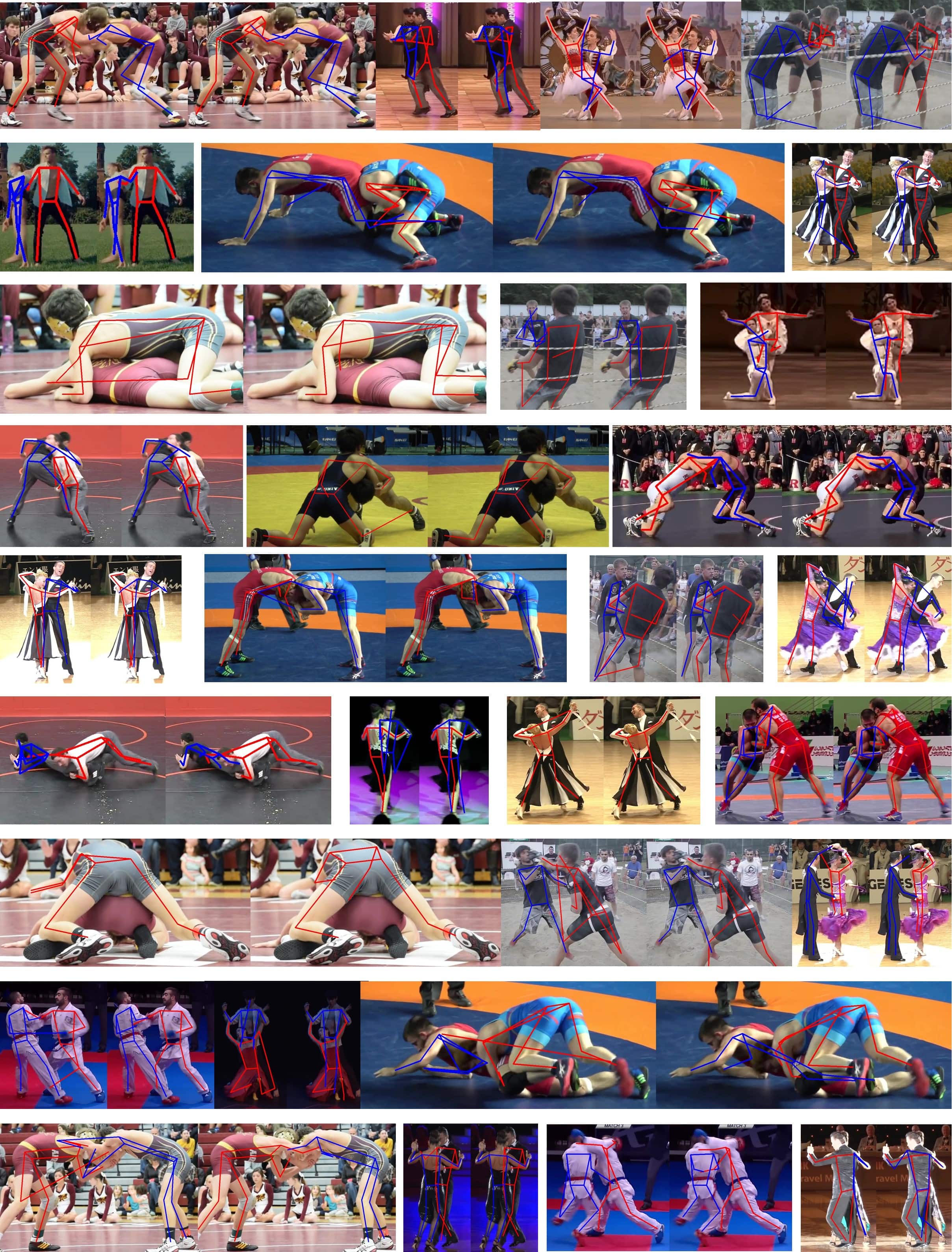}
	\end{center}
	\caption{Result Gallery of the OCPose. For each example, the left result is from AlphaPose+ while the right is from ours OPEC-Net.}
	\label{fig:RG-OCPose}
\end{figure*}

\newpage
\begin{figure*}[!h]
	\begin{center}
		\includegraphics[width=0.95\linewidth]{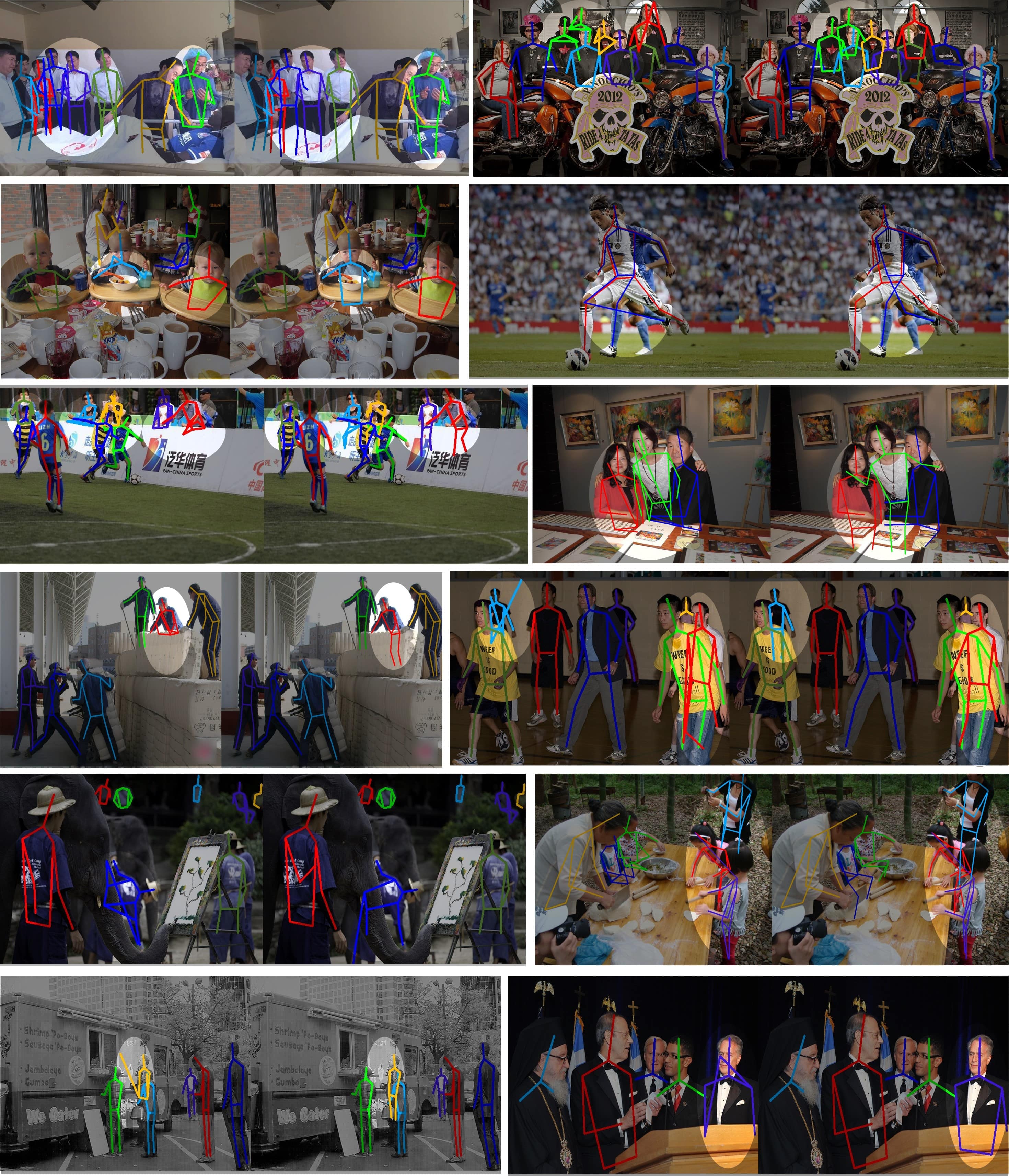}
	\end{center}
	\caption{Result Gallery of the CrowdPose. For each example, the left result is from AlphaPose+ while the right is from ours OPEC-Net.}
	\label{fig:RG-CrowdPose}
\end{figure*}

\newpage
\begin{figure*}[!h]
	\begin{center}
		\includegraphics[width=0.95\linewidth]{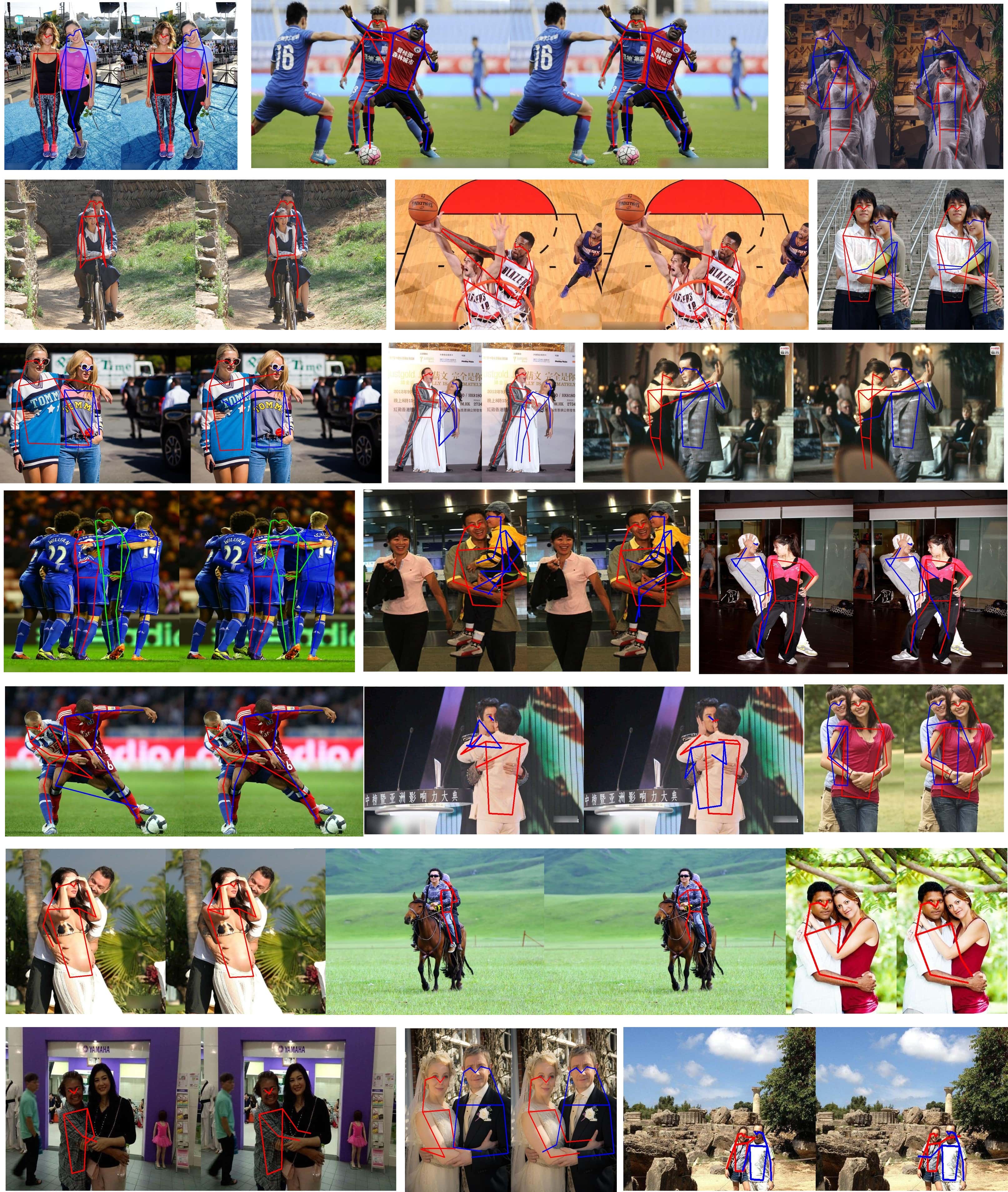}
	\end{center}
	\caption{Result Gallery of the OCHuman. For each example, the left result is from AlphaPose+ while the right is from ours OPEC-Net.}
	\label{fig:RG-CrowdPose}
\end{figure*}

\end{document}